\documentclass[10pt,twocolumn,letterpaper]{article}

\usepackage{iccv}
\usepackage{graphicx}
\usepackage{amsmath}
\usepackage{amssymb}
\usepackage{boldline}
\usepackage[ruled]{algorithm2e}
\usepackage{subfigure}

\def\SOTA{state-of-the-art\xspace}

\def\SAUG{SDA\xspace}

\usepackage{cite,nicefrac}
\usepackage{listings}

\usepackage{enumitem}
\setitemize{noitemsep,topsep=3pt,parsep=0pt,partopsep=0pt}

\usepackage[margin=4pt,font=small,labelfont=bf,labelsep=endash,tableposition=top]{caption}

\usepackage{xcolor}
\definecolor{codegreen}{rgb}{0.5,0.5,1}
\definecolor{codeblue}{rgb}{0.5,0.5,1}
\definecolor{codegray}{rgb}{0.6,0.6,0.6}

\usepackage[pagebackref=true,breaklinks=true,letterpaper=true,colorlinks,bookmarks=false]{hyperref}

\iccvfinalcopy %

\def\iccvPaperID{6658} %

\begin{document}

\title{A Simple Baseline for Semi-supervised Semantic Segmentation 
\\ 
with  Strong Data Augmentation\thanks{JY and YL contributed equally to this work.
\it Accepted to Proc.\ Int.\ Conf.\ Computer Vision (ICCV) 2021.
}
}

\author{
Jianlong Yuan$^1$,
~
~
~
~
Yifan Liu$^2$, 
~
~
~
~
Chunhua Shen$^2$, 
~
~
~
~
 Zhibin Wang$^1$,
~
~
~
~
Hao Li$^1$
\\[.1cm]
$^1$  Alibaba DAMO Academy  %
~~~~~
$^2$ The University of Adelaide, Australia
}

\maketitle

\begin{abstract}
Recently, significant progress has been made on semantic segmentation. However, the success of supervised semantic segmentation typically relies on a large amount of labeled data, which is time-consuming and costly to obtain. Inspired by the success of semi-supervised learning methods %
for 
image classification, here we propose a simple yet effective semi-supervised learning framework for semantic segmentation. We demonstrate that the devil is in the details: a set of simple designs and training techniques can collectively improve the performance of semi-supervised semantic segmentation significantly. Previous works \cite{chen2005naive,ouali2020semi} fail to 
effectively 
employ strong augmentation in pseudo-label learning, as the large distribution %
disparity 
caused by strong augmentation harms the batch normalization statistics. We design a new batch normalization, namely distribution-specific batch normalization (DSBN) to address this problem and 
show 
the importance of strong augmentation for semantic segmentation. Moreover, we design a self-correction loss, which is effective in terms of noise resistance. We conduct a series of ablation studies to show the effectiveness of each component. Our method achieves state-of-the-art results in the semi-supervised settings on the Cityscapes 
and Pascal VOC datasets.
\end{abstract}

\section{Introduction}
\begin{figure}
    \centering
     \subfigure[mean]{\includegraphics[scale=0.45]{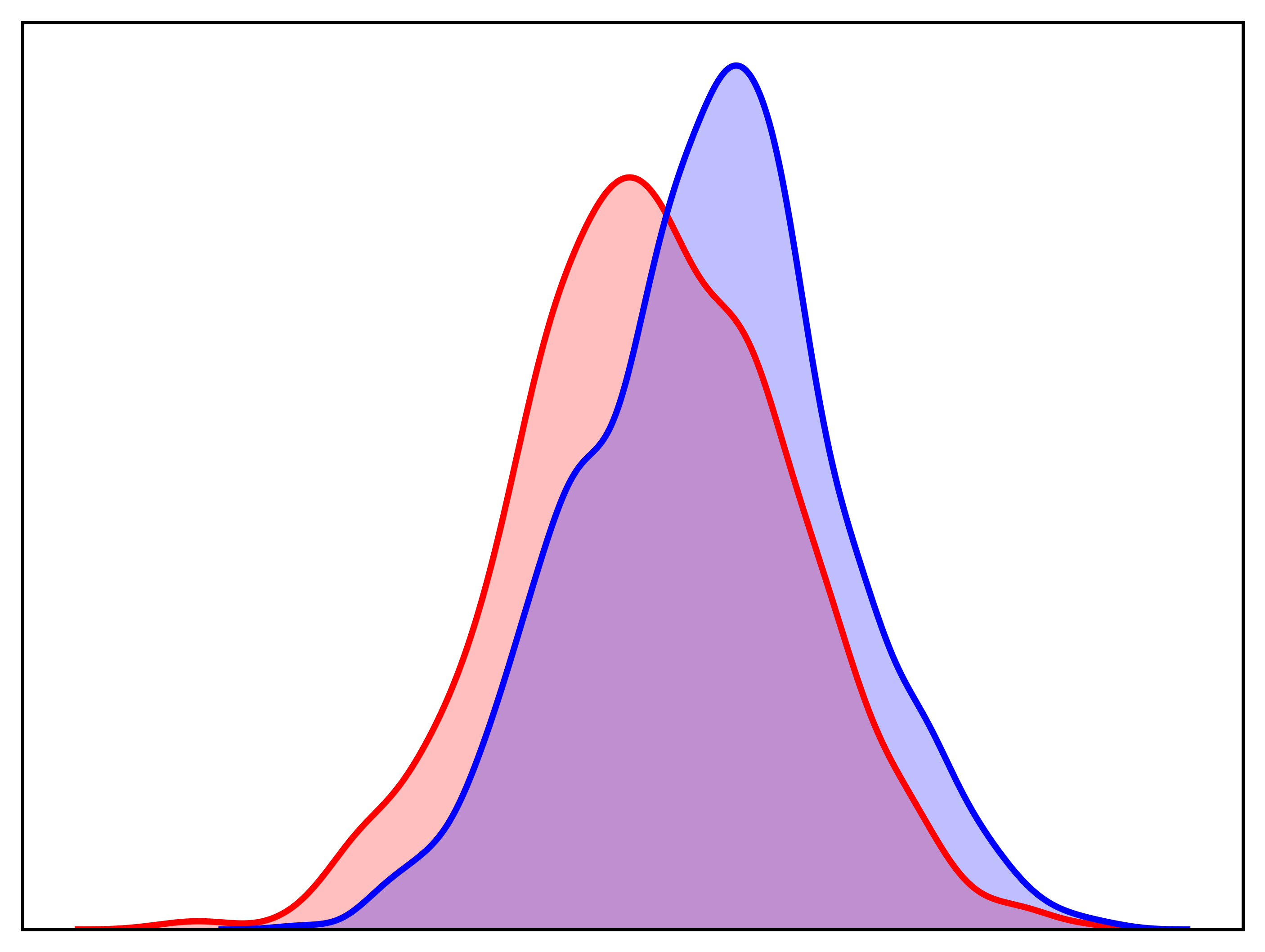}}
     \subfigure[variance]{\includegraphics[scale=0.45]{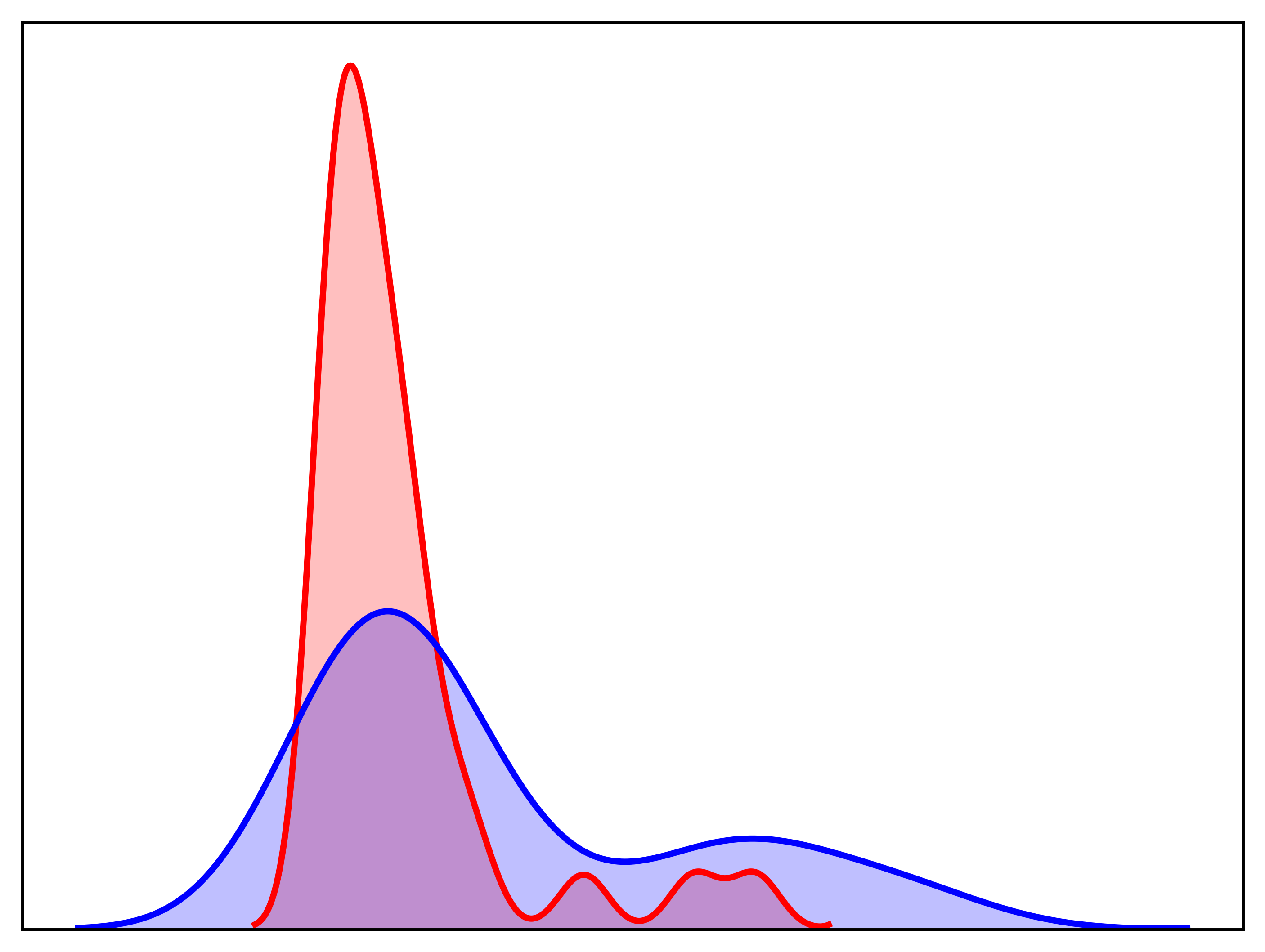}}
    \vspace{-0.4cm} 
    \caption{Distribution %
    mismatch 
    of BN %
    statistics 
    between weak %
    augmentation and strong augmentation.
    Red and blue lines denote BN statistics trained with 
    weak and strong data augmentation, respectively. 
}
    \label{fig:bn dif}
\end{figure}

Semantic segmentation/pixel labelling is one of the core %
tasks 
in visual understanding,
which is widely used in %
scene parsing, human body parsing, and many downstream applications.
It is a per-pixel classification problem, which classifies each pixel in the image into a predefined set of categories.
In the past few years, 
semantic segmentation methods based on deep convolution neural networks (CNNs) have made tremendous progress \cite{%
cityscpaes, %
everingham2012pascal%
}. 
Note that, the common prerequisite for all these successes is the
availability of a massive amount of pixel-level 
labeled data.
Unfortunately, labeling such %
datasets 
is
very
expensive and time-consuming, especially in dense prediction problems, such as semantic segmentation. 
As we need to label each pixel, %
it can be 
$60$ times more %
expensive 
than 
image-level annotation. 

Recent research reveals that semi-supervised learning (SSL), which uses a large amount of unlabeled data %
together 
with a small amount of labeled data, is greatly beneficial to classification \cite{xie2020self, berthelot2019mixmatch, sohn2020fixmatch, berthelot2019remixmatch, 
chen2020big, chen2020improved, grill2020bootstrap
}. These methods can be %
grouped 
into consistency methods \cite{berthelot2019mixmatch, sohn2020fixmatch, berthelot2019remixmatch}, pseudo-labelling  methods \cite{xie2020self}, representation learning \cite{%
grill2020bootstrap}.
A few works %
attempted 
to apply SSL to semantic segmentation. The naive student \cite{chen2005naive} uses a large model to generate the pseudo-labels with unlabeled video sequences, and apply iterative training for further improvement. They only consider the pseudo-labels with the original %
input 
images without perturbations. %
Recent work %
of 
\cite{ouali2020semi} considers  adding perturbations to the %
images for semantic segmentation.
They forward images with different perturbations in different sub-decoders and enforce the consistency between sub-decoders %
and 
the main decoder.
Other 
work pays attention to generative methods \cite{hung2018adversarial, mittal2019semi, souly2017semi}. AdvSemiSeg %
\cite{hung2018adversarial} and %
the work in \cite{mittal2019semi} both %
employ 
Generative Adversarial Network (GAN) and train the model with a discriminative loss over unlabeled data and a supervised loss over labeled data. 
Here, 
we propose an effective and efficient framework to apply SSL to semantic segmentation. We 
employ  
strong augmentation to make %
better 
use of the unlabeled data. 
Previous works based on consistency learning~\cite{chen2020improved,xie2019unsupervised,berthelot2019remixmatch} have shown that adding noise in the procedure of learning, the pseudo-labels %
help improve the performance in image classification. %
Motived by those works, we propose to apply strong 
augmentation to semantic segmentation. 
However, strong augmentation %
would inevitably 
affect the distribution of mean and variance in the batch normalization (BN) as shown in Figure~\ref{fig:bn dif}. 
As a consequence, \textit{one needs to be careful with computing the BN statistics in this scenario}. 
This may be the reason why %
strong augmentation was not %
used in SSL 
semantic segmentation \cite{chen2005naive}.
Instead, 
multiple branch networks were employed to process different perturbations, as in \cite{ouali2020semi}.  In order to avoid the \textit{distribution shift} caused by strong data augmentations, we propose a simple %
yet 
effective method, 
namely, the 
distribution-specific batch normalization (DSBN). We forward the strongly  augmented data and the weakly-augmented data with different batch %
statistics 
during training, and inference the model with the batch %
statistics
calculated by the weakly-augmented data.

Furthermore, as the teacher network may not perform well under all situations, some unreliable regions may be included in the generated 
pseudo-labels. Thus,  directly learning from all the pixels %
will inevitably  
introduce label noise. Inspired by previous work in \cite{wang2019symmetric}, which selects the unreliable image-pairs, and exchanges the prediction with the learning target to %
suppress the negative impact of  label 
noise, we design a new self-correction loss ({SCL}) and dynamically modify the weight and learning target for each pixel for semantic segmentation.

Our approach achieves state-of-the-art performance on the Cityscapes %
\cite{cityscpaes} and Pascal VOC datasets \cite{everingham2012pascal} under a semi-supervised setting.
Our main contributions are summarized as follows.
\begin{itemize}
\itemsep 0cm 
  \item We propose an effective and efficient semi-supervised learning framework for semantic segmentation. We 
  employ 
  strong augmentation during training
  without 
  modifying the network structure, such as 
  introducing sub-networks. 
  \item  A distribution-specific batch normalization is proposed to 
  accommodate 
  the 
  discrepancy 
  of data distribution in a batch caused by strong data augmentation. Besides, we design a self-training robust loss to %
  alleviate the negative impact of label noise. 
  The loss function %
  is able to 
  correct the noisy label to some extent, %
  thus
  resisting noise.
  \item We empirically demonstrate the effectiveness of our approach, 
  including a comparison with state-of-the-art methods,
  and an in-depth analysis of our approach with a detailed ablation study in the semi-supervised setting.
\end{itemize}

Next, we review some work relevant to ours.

\noindent\textbf{Semi-supervised classification.} 
Most %
previous works
on SSL focus on classification tasks.
Current state-of-the-art representative methods include consistency based methods \cite{berthelot2019mixmatch, sohn2020fixmatch, berthelot2019remixmatch}, Pseudo labeling \cite{xie2020self} and representation learning methods \cite{chen2020big, chen2020improved, grill2020bootstrap}.
Consistency based methods %
exploit the fact that 
the prediction of an unlabeled image should not change significantly with %
minor 
perturbations. UDA \cite{xie2019unsupervised} and ReMixMatch \cite{berthelot2019remixmatch} both use a weakly-augmented example to generate an artificial label and enforce consistency against strongly-augmented examples. FixMatch implements consistency training by applying strong (CutOut \cite{devries2017improved}, CTAugment \cite{berthelot2019remixmatch}, and RandAugment \cite{cubuk2020randaugment}) and weak %
data augmentation to the inputs in both the unlabeled loss and the pseudo label generation, thus encouraging the outputs to be consistent for both augmented inputs. %
Pseudo labeling methods %
rely on the assumption that 
the pseudo-labels generated by a %
teacher 
model can benefit the training of new models. Noisy student training \cite{xie2020self} is an iterative self-training method %
in this category.  
The correspondence among different transformations is employed for representation learning. %
Then a small amount of data is used to fine-tune the model.

\noindent\textbf{Semi-supervised  semantic segmentation}.  Early methods %
use 
a GAN model in semi-supervised segmentation %
\cite{hung2018adversarial, mittal2019semi, souly2017semi}, where an adversarial loss is trained on unlabeled data. Consistency regularization based approaches \cite{ouali2020semi, french2019semi, ke2020guided, luo2020semi, luo2020semi} have also been proposed recently, and achieve promising results. They depend on the fact that unlabeled data and labeled data have the same distribution, so it expects the trained model to have consistent and reliable predictions for both the unlabeled data and the labeled data. In these works, different unlabeled images under different transformations are used as the input %
to 
the model, and consistency loss is enforced on the prediction mask of the model. %
To date, 
self-training pushes the performance of the 
state-of-the-art %
\cite{chen2005naive, zoph2020rethinking}. %
Note that none of them attempted 
to solve the %
negative 
impact caused by noisy pseudo-labels.

\noindent\textbf{Data augmentation.}
Data augmentation is an effective regularization technique. The basic policy of data augmentation contains random flip, random crop, etc., which are commonly used 
in training vision models.
In addition,
the inception preprocess \cite{szegedy2015going} is more %
sophisticated 
when the input image color is randomly disturbed. Recently, AutoAugment \cite{cubuk2018autoaugment} improves the inception preprocess for image classification, using reinforcement learning to search 
for 
the optimal combination of %
augmentation 
strategies. RandAugment \cite{cubuk2020randaugment} proposes a significantly reduced search space which allows it to be trained on the target task and dataset, %
thus removing the need for 
a separate proxy task or dataset. 
All of these works 
tackle
image classification. In this paper, we %
propose 
to apply strong augmentation to semantic segmentation.  

\noindent \textbf{%
Separated 
batch normalization layers.}
It was observed in the literature that 
employment of a separated batch normalization to process out-of-distribution samples can lead to performance improvement. 
AdvProp~\cite{xie2020adversarial} uses an auxiliary batch normalization  for accommodating domain shift coming from adversarial examples, which are proposed for more effective adversarial training in the context of 
supervised image-level  classification.
We are inspired by this approach but employ an auxiliary batch normalization to process strongly augmented training samples which exhibit different statistics from that of the weakly augmented samples. 
Split-BN~\cite{zajkac2019split}, which shares $\beta$
and $\gamma$ parameters, is proposed for class mismatch and image
distortions of unlabeled datasets. TransNorm~\cite{wang2019transferable} is %
designed 
for unsupervised domain adaptation. %
TransNorm calculates $\alpha$ for a different domain. %
Here, 
our motivation is to solve the negative impact %
of BN statistics discrepancy 
caused by 
strong augmentation %
in image segmentation tasks.

\noindent\textbf{%
Robust 
loss functions for training with noisy data.}
A few 
works pay attention to the loss function %
for learning with noisy labels, and 
achieve  improved results. The generalized cross-entropy \cite{zhang2018generalized} was proposed to achieve the advantages of both mean absolute error (MAE) and categorical cross-entropy losses (CCE). 
Inspired by the symmetry of the Kullback-Leibler divergence, the symmetric cross-entropy \cite{wang2019symmetric} was proposed by combining a noise tolerance term, namely reverse cross-entropy loss, with the standard CCE loss. Nevertheless, since semantic segmentation is a pixel-level classification task,
directly applying these methods to segmentation, 
which are designed for image-level classification,  
is unlikely to yield satisfactory results.

\section{Our Method}
\label{sec:methods}

\subsection{Semi-supervised  Semantic Segmentation}
In this section, we introduce our simple semi-supervised learning framework for semantic segmentation. Different from CCT \cite{ouali2020semi}, which needs to design auxiliary decoders for different perturbations, our method can be applied to any existing segmentation network, as %
shown 
in Algorithm~\ref{algorithm}. 

Given a small set of labeled training examples and a large set of unlabeled training examples, the small set of labeled training examples are exploited to train an initial  teacher model %
using the 
standard cross-entropy loss. We then use the teacher model to generate pseudo-labels on 
the 
unlabeled images with a test-time augmentation. 
Following the Naive student method \cite{chen2005naive}, we only generate hard pseudo-labels, \textit{i.e.}, one-hot vectors. Then, we train a student model with \textit{strong augmentation} to make %
better 
use of the unlabeled data. Thus, the data distribution in a batch  %
is 
disturbed by the strong augmentation process, 
causing mismatch to that of the samples processed by weak/standard augmentation.  Here, 
we %
propose 
the distribution-specific batch normalization (DSBN) 
to alleviate the negative impact of this batch distribution 
mismatch. 
Moreover, motivated by the
symmetric cross entropy \cite{wang2019symmetric}, 
in order to learn with potentially noisy 
labels, we develop a new loss function, namely, self-correction loss ({SCL}). {SCL} is used to 
accommodate learning with noisy labels.
Finally, %
as in 
\cite{chen2005naive}, we iterate the process by putting back the student as a teacher to generate new pseudo-labels and train a new student.

\begin{algorithm}[t]
\caption{\small  Semi-supervised Learning Framework}
\label{algorithm} 
\footnotesize 
\LinesNumbered
 \textbf{Labeled images:} $n$ pairs of images $x_{i}$ and corresponding labels $y_{i}$\\
\textbf{Unlabeled images:} $m$ images without labels \{($\widetilde{x}_{1}$, $\widetilde{x}_{2}$, ..., $\widetilde{x}_{m}$\} \\
\textbf{Step1:} Optimize the teacher model with the cross-entropy loss on labeled images;\\

\While{iteration \textless %
maximum number of iterations 
}{

\textbf{Step2:} Use the teacher model to generate hard pseudo-labels (one-hot encodings) for clean 
(\textit{i.e.}, not distorted) unlabeled images with multi-scale and flip inference to get $m$ pairs of images with pseudo-labels \{($\widetilde{x}_{1}$, $\widetilde{y}_{1}$), ($\widetilde{x}_{2}$, $\widetilde{y}_{2}$), ..., ($\widetilde{x}_{m}$, $\widetilde{y}_{m}$)\} \\

\For{$epoch=1 ,..., n_{epoch}$ }{
	$\hat{x_i}$ $=$ \SAUG($\widetilde{x_i}$), $\forall i=1,2, ..., m;$

[$y^{*}$, $\widetilde{y}^{*}$] $=$ $f$([$x$, $\hat{x}$]);\\
loss $=$ CE($y^{*}$, $y$) $+$ SCL($\widetilde{y}^{*}$, $\widetilde{y}$); \\
minimizing the loss and update student parameters $\theta_{s}$; \\
}
Re-initialize the teacher parameters $\theta_{t}=\theta_{s}$;\\
Jump to \textbf{step 2}.
}
\end{algorithm}

%
%
%
%
%
%
%
%
%
\lstset{
  backgroundcolor=\color{white},
  basicstyle=\fontsize{8.5pt}{9.5pt}\fontfamily{lmtt}\selectfont,
  columns=fullflexible,
  breaklines=true,
  captionpos=b,
  commentstyle=\fontsize{9pt}{10pt}\color{codegray},
  keywordstyle=\fontsize{9pt}{10pt}\color{codegreen},
  stringstyle=\fontsize{9pt}{10pt}\color{codeblue},
  frame=tb,
  otherkeywords = {self},
}
\begin{figure}[!b]
\begin{lstlisting}[language=python]
Transforms = ['contrast gamma', 'contrast linear', 'brightness', 'brightness channel', 'equalize', 'hsv', 'invert channel', 'blur','noise gau', 'noise pos', 'Channel shuffle', 'dropout', 'coarse dropout', 'multiply', 'salt pepper', 'solarize', 'jpeg compression']

Basic_transforms = ['random scale', 'random flip', 'random crop', 'Normalization']

# data augmentation func
def SDA(N):   
""" Generate a set of distortions.
    Args: 
    N: Num. of augment. transform. to apply sequentially
"""
    sampled_ops = np.random.choice(Transforms, N)
    return sampled_ops + Basic_transforms
\end{lstlisting}
\caption{Python code for Strong Data Augmentation %
in Numpy.}
\label{fig: strong aug}
\end{figure}

\vspace{0.2cm}
\noindent\textbf{Strong data augmentation for semantic segmentation (\SAUG).}
While %
augmentation strategies  for supervised and semi-supervised image classification %
were 
extensively studied \cite{berthelot2019remixmatch,cubuk2018autoaugment,xie2019unsupervised, sohn2020fixmatch}, 
much less 
effort has been made %
for semantic segmentation. 
We %
apply random augmentation  
to semantic segmentation.
As illustrated in Figure \ref{fig: strong aug}, we build a pool of operations, which contains $16$ image transformation methods. 
At 
each training iteration, we randomly select a series of operations from the pool, and then combine them with standard transformations of semantic segmentation (random scale, random crop, random flip, \textit{etc}.).

\begin{figure}[t]
  \centering
    \includegraphics[scale=0.6]{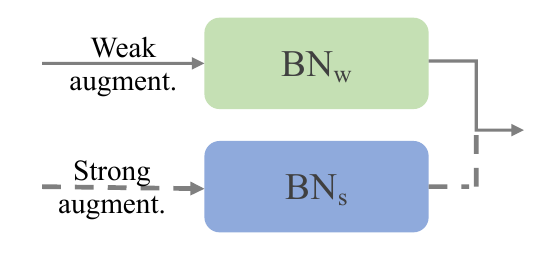}
  \caption{Distribution-Specific Batch Normalization ({DSBN}).  During the training phase, images with strong augmentation are 
  passed through BN$_s$, while images with weak augmentation are passed through standard BN$_w$. During the test phase, BN$_s$ is  discarded.
}
	\label{fig:DSBN}
\end{figure}

\subsection{Distribution-Specific Batch Normalization}

As mentioned in some previous works \cite{rota2018place, zhao2017pyramid, chen2018encoder}, batch normalization (BN) is crucial for semantic segmentation. BN  computes the mini-batch mean ($\mu_B$) and standard deviation ($\sigma_B$) \cite{ioffe2015batch}, formulated in Equation \eqref{equ: batch mean var}. Then, as shown in Equation \eqref{equ: batch norma}, BN %
uses a learnable scale ($\gamma$) and shift ($\beta$) parameter to transform the normalized distribution. Meanwhile, moving mean and moving variance are updated by Equation \eqref{equ: batch norm decay}. In the testing phase, BN uses the moving mean and moving variance for whitening input activations.
However, strong augmentations %
are likely to 
shift 
the distribution of natural images and lead to a domain gap between training images and testing images.
Thus, the moving mean and moving variance calculated during the training phase %
would 
harm the testing performance. To deal with %
this issue, 
we propose the distribution-specific batch normalization (DSBN). 
Specifically, 
we have two separate BNs: BN$_w$ for samples
of weak/standard augmentation; and  BN$_s$ for samples of strong 
augmentation  (see Figure~\ref{fig:DSBN}). In the training phase, if a data sample comes from pseudo-labels, then
{BN$_s$} is used in the forward; otherwise, BN$_w$ is used. In the testing phase, the {BN$_s$} is deprecated, and only the  BN$_w$ is used for normalization: 
\begin{equation}
    \mu_B = \tfrac{1}{m}\sum_{i=1}^{m}x_i ; \;\;   \sigma_B = 
    \Bigl
    [ {\tfrac{1}{m}\sum_{ i =1 }^{m}(x_i-\mu_B)^2 + \epsilon} \Bigr ] ^{0.5},
    \label{equ: batch mean var}
\end{equation}
\begin{equation}
    y_i = \gamma \frac{(x_i - \mu_B)}{\sigma_B} + \beta,
    \label{equ: batch norma}
\end{equation}
\begin{equation}
	\begin{split}
    \mu_{t+1} = \alpha \mu_{t} + (1 - \alpha) \mu_{t+1}, \\
    \sigma_{t+1} = \alpha \sigma_{t} + (1 - \alpha) \sigma_{t+1}.
    \end{split}
    \label{equ: batch norm decay}
\end{equation}

\subsection{Self-correction Loss} 

Previous work~\cite{wang2019symmetric} has shown that cross-entropy loss masks the model overfit to noisy labels on some easier classes and under learning on some harder classes in image classification.
As semantic segmentation is a per-pixel classification problem, similar challenges also exist. Moreover, in the semi-supervised learning framework, the learning targets are pseudo-labels, which may contain unreliable regions caused by the limited capacity of the current teacher model. 

To avoid overfitting to the noise contained in the pseudo-labels, we propose a Self-Correction Loss (SCL) for semantic segmentation. It assigns an adaptive weight by the confidence of the network output to each pixel during the training phase. Besides, we apply a noise-robust term Reverse Cross Entropy~\cite{wang2019symmetric} to the unreliable regions. Different from previous work, which chooses the reversed learning pairs by fixed confidence from human heuristics, we dynamically change the learning targets by comparing the output of the student network and the teacher network.

As shown in Equation~\eqref{equ: ASCL equation} ($i$ is the location index), $\mathbf{y}_i^*$ denotes the prediction of a pixel, $\mathbf{y}_i$ denotes the pseudo-label generated by the teacher, and $w_i$ represents the dynamic weight which is the largest activation after softmax among all $c$ classes %
(refer to Equation~\eqref{equ: ASCL equation weights}).  We use the prediction to automatically adjust the confidence during training. If the confidence is very high, 
we adopt positive learning. %
If the confidence is very low, we %
believe 
that the pseudo-label for this pixel is unreliable. Thus, reverse learning is applied. In addition, SCL is only applied to pseudo-labels:
\def\y{ {\mathbf y }} 
\begin{equation}
	\ell_i = w_i \cdot  {\y_i} \log( {\y_i^*})+(1-w_i) \cdot 
	{\y_i^*} \log( {\y_i}),
	\label{equ: ASCL equation}
\end{equation}
\begin{equation}
\label{equ: ASCL equation weights}
	w_{i} = \max(\frac{ \exp(y_{i0}^*)}{\sum_{j=0}^{c} \exp(y_{ij}^*)} , ... ,\frac{ \exp(y_{ic}^*)}{\sum_{j=0}^{c} \exp(y_{ij}^*)}).
\end{equation}

\section{Experiments}
In this section, we first report the implementation details. Then we perform a series of ablation experiments, and analyze the results in detail. Finally, we report our results compared with other state-of-the-art methods.
\subsection{Implementation Details}

\noindent\textbf{Training.} Our implementation is built on Pytorch \cite{paszke2019pytorch}.
Following previous work \cite{chen2018encoder}, we use the ``poly" learning rate policy, where the base learning rate is multiplied by $(1-%
{ iter}/{ 
{ iter}_{\max}})^{ power}$. The initial learning rate is set to $0.01$, and power is set to $0.9$. We train the network using mini-bath stochastic gradient descent (SGD).  
We employ the crop size of $769 \times 769$ on Cityscapes with DeepLabv3Plus. More details is same as \cite{chen2018encoder}.
In the addition study, following \cite{zhao2017pyramid, yuan2020multi, chen2018encoder, lin2017refinenet, wu2019wider}, we adopt the mean intersection of union (mIoU) as the evaluation metrics.

\noindent\textbf{Data augmentation.} Following \cite{chen2018encoder}, we use mean subtraction, and apply data augmentation by random resize between 0.5 and 2 and random left-right mirror during initialization training. %
In the semi-supervised learning phrase, we use strong augmentation for improving the performance of self-training.

\noindent\textbf{Datasets.} Following \cite{chen2005naive}, We conduct the main experiments and ablation studies on the Cityscapes dataset \cite{cityscpaes}. This large-scale data set contains different stereo video sequences recorded in $50$ different urban street scenes. In addition to $20$k weak annotation frames, there are $5,000$ high-quality pixel level annotation frames, in which, $2,975$; 
$500$ and $1,525$ were used for training, validation and testing. In addition, there are $20$k images with rough annotations, namely train-coarse. We randomly subsample %
$\nicefrac{1}{8}$, 
$\nicefrac{1}{4}$, and 
$\nicefrac{1}{2}$
of images in the standard training set to construct the pixel-level labeled data. The remaining images in the training set are used as unlabeled data. 

At last, we report our results on the Pascal VOC 2012 dataset \cite{everingham2012pascal}, which contains 21 classes including background. The standard Pascal VOC 2012 dataset has $1,449$ images as the training set and $1,456$ images as the validation set. We construct $1,449$ images as the pixel-level labeled data. The images in the augmented set \cite{SBD} (around 9k images), are used as unlabeled data.

\begin{table*}[tb]
\footnotesize 
\centering 
\begin{tabular}{l c c c c c c c c c}
\hlineB{2}
Model &s.t. &\SAUG & DSBN &SCL &iter.\  & mIoU-$\nicefrac{1}{8}$ (\%) &mIoU-$\nicefrac{1}{4}$ (\%) & mIoU-$\nicefrac{1}{2}$ (\%) & mIoU-Full (\%) \\
\hline\hline
Deeplabv3plus     & && & & & 68.9&73.4&76.9&78.7 \\
Deeplabv3plus     &$\checkmark$ && & & &71.2&76.2&78.0&79.5 \\
Deeplabv3plus     &$\checkmark$ &$\checkmark$& & & &71.3 &76.3 &78.2 &79.7 \\
Deeplabv3plus     &$\checkmark$ &$\checkmark$&$\checkmark$ & & &72.3&76.9&78.7&80.0 \\
Deeplabv3plus     &$\checkmark$ &$\checkmark$ &$\checkmark$&$\checkmark$ & & 72.8&77.4&78.7&80.4\\
Deeplabv3plus     &$\checkmark$ &$\checkmark$ &$\checkmark$&$\checkmark$ &$\checkmark$ & 74.1&77.8&78.7&80.5\\
\hlineB{2}
\end{tabular}
\caption{Ablation study on the proposed semi-supervised learning framework. The %
model here 
is 
Deeplabv3plus with Xception65 backbone. `s.t.' denotes self-training with pseudo-labels without strong augmentation. `\SAUG ' means the strong data augmentation. `DSBN' means the distribution specify batch normalization. `SCL' is the proposed self correction loss. `iter.' represents iterative training. The results are evaluated on the validation set with the single-scale input.  `mIoU-$\nicefrac{1}{n}$' means that we use  $\nicefrac{1}{n}$ labeled data and the remaining images in the training set are used as unlabeled data. `mIoU-Full' means that we use all the labeled data in the training set, and the train-coarse are employed as the unlabeled dataset.}
\label{tab: ablation for ist}
\end{table*}

\subsection{Ablation Study}
In this subsection, we conduct the experiments to explore the effectiveness of each proposed module under different semi-supervised settings. First, we establish the baseline for our experiments. We evaluate DeepLabv3Plus based on Xception65 with cross entropy on the validation set. Following \cite{chen2005naive, xie2020self}, self training adopts the baseline model as the teacher model and the teacher model generates pseudo labels on the unlabeled dataset, then the student model which is not smaller than the teacher model is trained on pseudo labels and original labeled images. We use the scales including $0.5$, $0.75$, $1.0$, $1.5$, $1.75$  and flip for the remaining images to generate pseudo labels. For fair comparison and quick training, the student model is the same model initialized by the teacher model. The main results are shown in Table~\ref{tab: ablation for ist}, and we will give detail discussions on each module in the following sections. All the ablation study are conducted on Cityscapes.

\subsubsection{Different semi-supervised settings}

We follow previous works~\cite{french2019semi,souly2017semi,hung2018adversarial} to divide the train set into labeled data and unlabeled data according to different proportions. We 
use
$\nicefrac{1}{8}$,
$\nicefrac{1}{4}$,
and $\nicefrac{1}{2}$
of the training dataset as labeled data and other images as the unlabeled data. The evaluation results are shown in Table~\ref{tab: ablation for ist} as 
mIoU-$\nicefrac{1}{8}$, mIoU-$\nicefrac{1}{4}$, mIoU-$\nicefrac{1}{2}$, respectively. From the Table \ref{tab: ablation for ist}, we can see that our algorithm can effectively improve the performance by $5.2\%$, $4.4\%$, $1.8\%$, $1.8\%$, respectively. 
When the total number of labeled data and unlabeled data does not change, the semi-supervised learning framework will have a larger improvement, if we have less labeled images. In the meantime, the final results will be better if we have more high-quality labeled images. Besides, we can see that only half the amount of labeled data with the help of proposed semi-supervised learning framework can achieve the same results (mIoU of $78.7\%$) as we use all labeled training data with supervised learning (row 1 and column 10).

When the total number of available data increases, semi-supervised learning will play a more valuable role. We  use all high-quality labeled data in the training set as the labeled data and employ the train-coarse as unlabeled data to conduct the experiment, and the evaluation results are referred to as mIoU-Full in Table~\ref{tab: ablation for ist}. We can see that under a strong baseline with large amount of labeled data, if extra unlabeled data can be obtained, the performance will be further improved with the proposed semi-supervised learning framework.

\begin{figure}[t]
  \centering
    \includegraphics[scale=0.8]{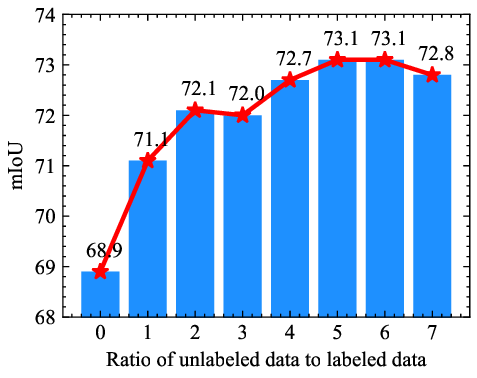}
  \caption{Different ratios of unlabeled data to labeled data. 
  }
	\label{fig:paper_Unlabel_images_multiples}
	\vspace{-1em}
\end{figure}

In practical applications, labeled data is often limited, but unlabeled data is easy to %
obtain.  
To explore the impact of increasing the ratio of unlabeled data in the training set.
We fix the labeled data as %
$ \nicefrac{1}{8}$
of the training set, and increase the number of the unlabeled data. The results are shown in Figure~\ref{fig:paper_Unlabel_images_multiples}.
From Figure~\ref{fig:paper_Unlabel_images_multiples}, we can see that increasing ratios of the unlabeled data to the labeled data would  improve the results, but the growth trend of the performance is gradually flattening limited by the capacity of the initial teacher model trained on the fixed labeled data. The performance %
even slightly drops if too many unlabeled images are introduced during training.

\subsubsection{Impact of the strong augmentation}

In this section, we will show the impact of the proposed strong augmentation and the DSBN. First, as shown in Table~\ref{tab: ablation for ist}, we compare the results of the naive self-training and the self-training combined with strong augmentation and the proposed DSBN. Although self-training can improve the performance over the baseline, 
self training directly uses pseudo labels that makes the network difficult to learn and introduces noise, as discussed in Section \ref{sec:methods}. Meanwhile, in the case of only using strong augmentation, there is no significant improvement in performance.
On the contrary, our method performs favorably against naive self training with strong augmentation and DSBN. In particular, compared with naive self training, our method improves the mIoU by $1.1\%$, $0.7\%$, $0.7\%$ and $0.5\%$ based on DeepLabv3Plus. In the following sections, we  analyze  the distribution of the BN statistics, and then we compare the strong augmentation with different BN settings. Finally, we provided some discussions on how does the strong augmentation work with ground truth labels and the pseudo labels.

\noindent\textbf{Visualization of the BN statistics.}
To verify our statement that the distribution of the BN will be affect by the strong augmentation, we show the statistics of BN training with strong augmentation input images and the weakly augmentation input images. The visualization results can be found in Figure~\ref{fig:bn dif}.

The distribution of the mean and variance obtained from the whole training statistics deviates from the test set after adding strong augmentation. As shown in Figure \ref{fig:bn dif}, we calculate the distribution of the mean and the variance of the BN with the weakly augmentation and the strong augmentation. The blue line denotes the distribution of the BN with weakly augmentation, while the red line represents the BN with strong augmentation. We can see that strong augmentation has changed the distribution of BN statistics, which may lead to a domain gap. We think this is the reason why strong augmentation has not brought much performance improvement, as shown in Table \ref{tab: ablation for ist}.

\noindent\textbf{Impact of the DSBN.}
To show the effectiveness of proposed DSBN, we conduct experiments to compare three different BN settings as follows:
\begin{itemize}
    \item Trainable BN: the mean and the variance are updated during training with both weak augmentation data and strong augmentation data.
    \item Fixed BN: the mean and the variance are fixed during training, and initialized with the pre-trained weights from the ImageNet classification.
    \item DSBN: different mean and the variance are updated during training for weak augmentation data and strong augmentation data, separately.
\end{itemize}
The baseline model is DeepLabV3Plus trained with $1/8$ labeled data on the training set. The other $7/8$ images are used as the unlabeled dataset in the semi-supervised learning. The results are shown in Table~\ref{tab: ablation for bn}. We can see that with the proposed DSBN, the performance can be improved by $0.7\%$, which indicates the DSBN can help to solve the negative effect caused by the strong augmentation.

\begin{table}[b]
\begin{center}
\resizebox{0.6\linewidth}{!}{
\begin{tabular}{l c }
\hlineB{2}
Baseline  & mIoU (\%)  \\
\hline\hline
ST     & 71.2 \\
ST + \SAUG  + Trainable BN &71.3 \\
ST + \SAUG  + Fixed BN   &71.6 \\
ST + \SAUG  + DSBN & 72.3 \\
\hlineB{2}
\end{tabular}
}
\end{center}
\caption{Ablation study for different batch normalization methods.}
\label{tab: ablation for bn}
\end{table}

\noindent\textbf{Discussions.}
To further understand the impact of the strong augmentation and DSBN on different data groups, we apply strong augmentation to only ground truth labels and pseudo labels.
\begin{itemize}
    \item \textit{$\nicefrac{1}{8}$ ground truth.} We train the model with only 1/8 images of the training dataset with corresponding ground truth labels.
    \item \textit{Full-ground truth.} We train the model with all images of the training dataset with corresponding ground truth labels.
    \item \textit{$\nicefrac{7}{8}$ pseudo labels.} We train the model with 7/8 images of the training dataset with corresponding pseudo labels.
\end{itemize}

For each settings, we apply weekly augmentation, strong augmentation and DSBN during training. The experiment results are shown in Table~\ref{tab: ablation for aug}. We can see that under the fully supervised settings, the strong augmentation can not bring extra improvements. Meanwhile, when applying to the unlabeled data with pseudo labels, the mIOU improved by $0.8\%$. These observations indicate that improvements under semi-supervised settings are not come from extra image transformation types on labeled data, but benefit from make further use of unlabeled data with strong augmentations.
From Table~\ref{tab: ablation for aug}, we also find that the DSBN can contribute to the training.
\textit{In addition, %
to our knowledge,  
this is the first time to apply the strong augmentation to semantic segmentation under fully-supervised setting, and %
improvement is observed.}

\begin{table}[t]
\begin{center}
\resizebox{0.8\linewidth}{!}{
\begin{tabular}{l |c }
\hlineB{2}

\hline
~~~ Methods  & mIoU (\%)\\
\hline
\multicolumn{2}{c}{Training with ground-truth labels}\\
\hline
$ %
\nicefrac{1}{8}
$-ground truth     & 68.9 \\
$ %
\nicefrac{1}{8}
$-ground truth + \SAUG  &68.6 \\
$ %
\nicefrac{1}{8}
$-ground truth + \SAUG  + DSBN & 69.5\\
\hline
Full-ground truth     & 78.7 \\
Full-ground truth + \SAUG  &78.7 \\
Full-ground truth + \SAUG  +  DSBN & 79.2\\
\hline
\multicolumn{2}{c}{Training with pseudo-labels}\\
\hline
$ %
\nicefrac{7}{8}$-pseudo labels    & 70.6 \\
$ %
\nicefrac{7}{8}
$-pseudo labels + \SAUG  &71.4 \\
$ %
\nicefrac{7}{8}
$-pseudo labels + \SAUG  + DSBN &72.5 \\
\hlineB{2}
\end{tabular}
}
\end{center}
\caption{Effect of the strong augmentation on different data.
We apply the strong augmentation to the baselines trained with only labeled data, and only unlabeled data with pseudo-labels, separately.}
\label{tab: ablation for aug}
\vspace{-1em}
\end{table}

\subsubsection{Impact of the self-correction loss}
To further improve the performance and reduce the impact of noise, we employ a self correction loss (SCL). From Table~\ref{tab: ablation for ist}, we can see that the proposed SCL will play a more important role when the pseudo labels contain more noise. The teacher model trained with a limited labeled dataset may get less useful knowledge, and can not generalize well on unlabeled dataset. Thus, the level of the noise is higher than that training on more labeled data. When we train with 
$\nicefrac{1}{8}$
or 
$\nicefrac{1}{4}$
labeled images, the SCL can improve the mIoU by 
$0.5\%$. 

\begin{figure}[th]
  \centering
    \includegraphics[scale=0.705]{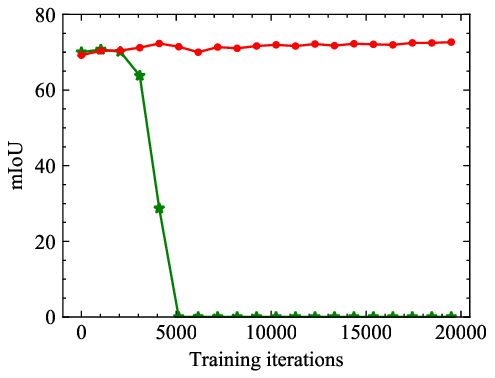}
  \caption{This curve %
  shows
  the 
  validation mIoU w.r.t.\ the number of iterations during training.
  The green line represents the SCE loss \cite{wang2019symmetric}. In addition, we set the same ratio between CE and RCE. The red line represents our proposed loss with the same initial weights as the SCE loss. }
	\label{fig:scl iterations}
	\vspace{0.2in}
\end{figure}

Furthermore, we show the mIoU during the training phase with and without the proposed SCL in Figure \ref{fig:scl iterations}. When training without the SCL, the result will be divergent quickly. On the contrary, the result is relatively stable. 
\subsubsection{Impact of 
iterative 
training
}
\begin{figure}
  \centering
    \includegraphics[scale=0.649]{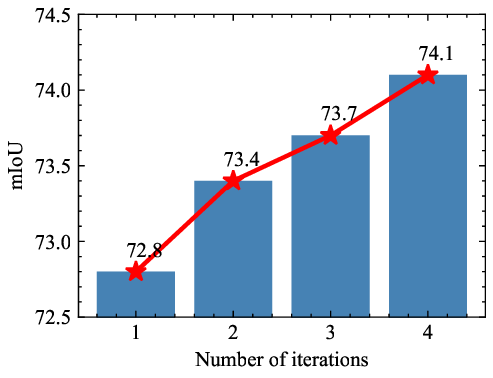}

  \caption{%
  The performance of the DeepLabV3Plus under different number of iterations.} 
	   \label{fig: iter_dp3}
 	\vspace{-0.2cm}
\end{figure}

Following Naive Students~\cite{chen2005naive}, we also test the impact of the 
iterative 
training in our semi-supervised learning framework. Table~\ref{tab: ablation for ist} has shown that the iteration training will become less useful if more labeled data can be obtained. Because the initial teacher model trained with more labeled data can already produce satisfactory pseudo labels. We show the change of the performance for the number of iterations in Figure~\ref{fig: iter_dp3}. The baseline model is DeepLabV3Plus trained with $\nicefrac{1}{8}$ labeled images. We can see that more iterations 
indeed 
help improve the performance, but the growth trend of the performance is gradually flattening.

\begin{figure*}
    \centering
    \includegraphics[width=0.97\linewidth]{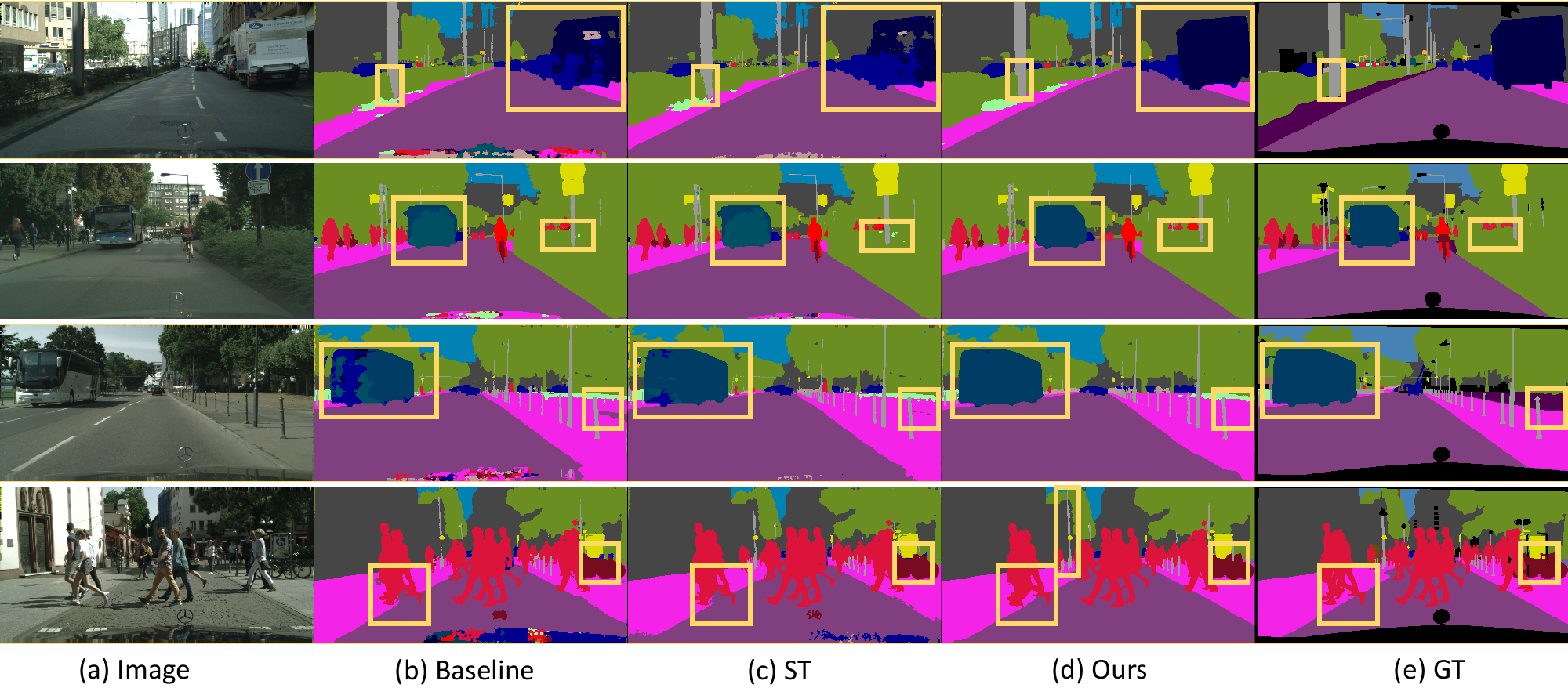}
    \caption{Qualitative results on the Cityscapes dataset.
    The baseline method is trained with 
$\nicefrac{1}{8}$  of all 
labeled images. 
     `ST' represents the self-training trained with 
$\nicefrac{7}{8}$
pseudo-labels without our methods. 
 `Ours'  %
 is 
 our framework with $\nicefrac{1}{8}$ labelled data.
    The proposed semi-supervised approach produces improved results compared to the baseline and the naive self training. We highlight the details in yellow boxes.}
    \label{fig: results}
    \vspace{-1em}
\end{figure*}

{\bf Visualization results}
Several visualization results are shown in Figure \ref{fig: results}. 
%
%
%
%
We can see  that the baseline and self-training can not well separate the objects (especially %
large-size 
objects such as  bus, truck, train, sidewalk) completely while ours corrects these errors. %
Our method also performs well on these small-size object classes, such as pole and bicycle, compared to the baseline model and self-training.

\subsection{Comparison with State-of-the-art Methods}
\noindent\textbf{Cityscapes.}
We conduct the comparison experiments with other state-of-the-art algorithms on the Table \ref{tab: SOTA  on different amount cityscapes}. To make a fair comparison, we also apply our proposed framework to DeepLabV2 following~\cite{hung2018adversarial}. In particular, DeepLabV2 is based on ResNet-101 initialized with the pre-trained weights on ImageNet classification. The proposed framework achieves 67.6\%, 69.3\%, 70.7\% with DeepLabV2 and 74.1\%, 77.8\%, 78.7\% with DeepLabV3Plus, respectively, which performs favorably against previous state-of-the-art methods. More details of ablation studies of with DeepLabV2 can be found in the supplementary materials.

\begin{table}
\begin{center}
\resizebox{\linewidth}{!}{
\begin{tabular}{r  c c c c c}
\hlineB{2}
Method &Model & $\nicefrac{1}{8} $ & $\nicefrac{1}{4} $ & $\nicefrac{1}{2} $ &Full\\
\hline\hline
AdvSemiSeg\cite{hung2018adversarial}&DeepLabV2 &58.8 & 62.3 &65.7 &66.0 \\
S4GAN + MLMT\cite{mittal2019semi} &DeepLabV2 &59.3 &61.9 &- &65.8 \\
CutMix\cite{french2019semi}&DeepLabV2 &60.3 &63.87 &- &67.7 \\
DST-CBC\cite{feng2020semi} &DeepLabV2 &60.5 &64.4 &- &66.9\\
ClassMix\cite{olsson2020classmix} &DeepLabV2 &61.4 &63.6 &66.3 &66.2 \\
ECS\cite{mendel2020semi} &DeepLabv3Plus &67.4 &70.7 &72.9 &74.8 \\
\hline
Baseline &DeepLabV2  &60.6 &66.7 &69.3 &70.1\\
Ours&DeepLabV2 &67.6 &69.3 &70.7 &70.1\\
Baseline& DeepLabV3Plus&68.9&73.4&76.9&78.7\\
Ours&DeepLabV3Plus &\textbf{74.1} &\textbf{77.8} &\textbf{78.7} &\textbf{78.7}\\
\hlineB{2}
\end{tabular}
}
\end{center}
\caption{Compared with  \SOTA methods on the Cityscapes val set. Here %
`$\nicefrac{1}{n}$'  means 
that 
we use  $ \nicefrac{1}{n} $ labeled data and the remaining images in the training set are used as unlabeled data.}
\label{tab: SOTA  on different amount cityscapes}
\vspace{-1em}
\end{table}

\noindent\textbf{Pascal VOC.}
We follow DeepLabV3Plus \cite{chen2018encoder} training details with the 513$\times$513 crop size on Pascal VOC \cite{everingham2012pascal}. Table \ref{tab: SOTA in voc} lists the performance results of other state-of-the-art methods and ours. We adopt the single-scale testing on our experiments. In order to make a fair comparison, we did experiments under different backbones. Our framework achieves 75.0\% mIoU based on Resnet-101\cite{He2016Deep} and 79.3\% mIoU based on Xception-65\cite{chen2018encoder}, which outperforms the PseudoSeg \cite{zou2020pseudoseg} by 1.8\%. At the same time, we find that our results based on Xception-65 can achieve the same performance 
as that of the fully supervised setting.
In addition, we find that the improvement of iterative training on Pascal VOC is limited, which might be due to the small number of samples.

\begin{table}
\begin{center}
\resizebox{0.7\linewidth}{!}{
\begin{tabular}{ r  c c c c c c c}
\hlineB{2}
Method     & Model &mIoU \\
\hline\hline
GANSeg\cite{souly2017semi} &VGG16 &64.1 \\
AdvSemSeg\cite{hung2018adversarial} &ResNet-101 &68.4 \\
CCT \cite{ouali2020semi} &ResNet-50 &69.4 \\
PseudoSeg \cite{zou2020pseudoseg} & ResNet-101 &73.2 \\
\hline
Ours &ResNet-101 &\textbf{75.0} \\
Ours &Xception-65 &\textbf{79.3} \\
Fully supervised &ResNet-101 &78.3 \\
Fully supervised &Xception-65 & 79.2 \\
\hlineB{2}
\end{tabular}
}
\end{center}
\caption{Comparison with  \SOTA on the Pascal VOC val set (w/ unlabeled data). We use the official training set (1.4k) as labeled data, and the augmented set (9k) as unlabeled data. 
}
\label{tab: SOTA in voc}
\vspace{-1em}
\end{table}

\section{Additional Ablation Study with DeepLabV2}

We conduct %
experiments to explore the effectiveness of each proposed module under \nicefrac{1}{8} labeled image settings. First, we establish the baseline for our experiments. We evaluate DeepLabV2 based on ResNet-101 %
on the validation set. Same as the experiments with the DeepLabV3Plus %
model,
self-training adopts the baseline model as the teacher model and the teacher model generates pseudo-labels on the unlabeled dataset, then the student model which is not smaller than the teacher model is trained on pseudo labels and original labeled images. We use the scales including $\{0.5, 0.75, 1.0, 1.5, 1.75\}$ and mirror for the remaining images to generate pseudo labels. For fair comparison,
the student model is the same model initialized by the teacher model. The main results are shown in Table~\ref{tab: ablation for ist of deeplabv2}.

\begin{table}[tb]
\begin{center}
\resizebox{\linewidth}{!}{
\begin{tabular}{l c c c c c c}
\hlineB{3}
Model &s.t.\  & \SAUG& DSBN &SCL &iter.\  & mIoU (\%) \\
\hline\hline
DeepLabV2     & & & & && 60.6\\
DeepLabV2     &$\checkmark$ & && & & 62.1 \\
DeepLabV2     &$\checkmark$ &$\checkmark$ && & &62.2 \\
DeepLabV2     &$\checkmark$ &$\checkmark$ &$\checkmark$ && & 63.8 \\
DeepLabV2     &$\checkmark$ &$\checkmark$ &$\checkmark$ &$\checkmark$& & 64.5 \\
DeepLabV2     &$\checkmark$ &$\checkmark$ &$\checkmark$ &$\checkmark$ &$\checkmark$& 67.6 \\
DeepLabV2-Full    & & & & && 70.1 \\
\hlineB{3}
\end{tabular}
 }
\end{center}
\caption{Ablation study on the proposed semi-supervised learning framework. Baselines are ResNet101-based Deeplabv2. 's.t.\' denotes self-training with pseudo labels without strong augmentation. \SAUG\  means the strong data augmentation. DSBN means the distribution specify batch normalization. SCL is the proposed self correction loss. `iter.' represents iterative training. The results are evaluated on the validation set with the single-scale input. Except %
the last row, 
all other experiments use $\nicefrac{1}{8}$ labeled data and the remaining images in the training set are used as unlabeled data. `$\cdot$ Full' means %
the fully supervised setting.}
\label{tab: ablation for ist of deeplabv2}
\end{table}

\begin{figure}
  \centering
    \includegraphics[scale=0.8]{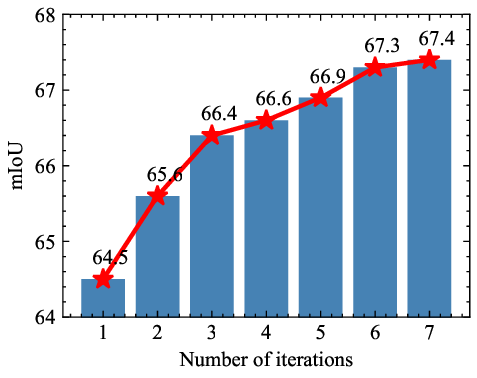}
  \caption{%
  Performance vs.\ number of iterations, using the DeepLabV2 model.}
	\label{fig:iterations for deeplabv2}
	\vspace{0.2in}
\end{figure}

We also report detailed performance under different iterations in Figure \ref{fig:iterations for deeplabv2}. We can see that more iterations %
in general 
help improve the performance, but the growth trend of the performance is gradually flattening.

\section{Conclusion}

In this work, we construct a simple semi-supervised learning framework for semantic segmentation. It employs strong augmentation with distribution-specific batch normalization, together with 
pseudo-label learning.
In particular, DSBN can effectively avoid the BN statistics shift caused by strong augmentation. Meanwhile, we have also designed a self-correction loss, %
which 
effectively %
alleviates the label noises introduced by pseudo labeling.
Quantitative and qualitative comparisons show that the proposed method performs favorably against recent state-of-the-art semi-supervised approaches.

{\small
\bibliographystyle{ieee_fullname}
\bibliography{FINAL}
}

\end{document}